\documentclass[letterpaper]{article} 
\usepackage{aaai23}  
\usepackage{times}  
\usepackage{helvet}  
\usepackage{courier}  
\usepackage[hyphens]{url}  
\usepackage{graphicx} 
\usepackage{natbib}  
\usepackage{caption} 
\usepackage{algorithm}
\usepackage{algorithmic}
\usepackage{amsmath}
\usepackage{multirow}
\usepackage[table,xcdraw]{xcolor}
\usepackage{amsfonts}
\usepackage{booktabs}
\usepackage{makecell}

\usepackage{newfloat}
\usepackage{listings}
\DeclareCaptionStyle{ruled}{labelfont=normalfont,labelsep=colon,strut=off} 
\floatstyle{ruled}
\newfloat{listing}{tb}{lst}{}
\floatname{listing}{Listing}
\usepackage[hang,flushmargin]{footmisc}

\title{SwiftAvatar: Efficient Auto-Creation of Parameterized Stylized Character \\ on  Arbitrary Avatar Engines}
\author{
    Shizun Wang\textsuperscript{\rm 1}\thanks{Corresponding Author},
    Weihong Zeng\textsuperscript{\rm 2*},
    Xu Wang\textsuperscript{\rm 2},
    Hao Yang\textsuperscript{\rm 2},
    Li Chen\textsuperscript{\rm 2},\\
    Yi Yuan\textsuperscript{\rm 2},
    Yunzhao Zeng\textsuperscript{\rm 2},
    Min Zheng\textsuperscript{\rm 2}
    Chuang Zhang\textsuperscript{\rm 1},
    Ming Wu\textsuperscript{\rm 1}\thanks{Corresponding Author},
}
\affiliations{
    \textsuperscript{\rm 1}Beijing University of Posts and Telecommunications \quad
    \textsuperscript{\rm 2}Douyin Vision\\
    
    \{wangshizun, zhangchuang, wuming\}@bupt.edu.cn \\
    \{zengweihong, wangxu.ailab, yang.hao, chenli.phd, yuanyi.cv, zengyunzhao, zhengmin.666\}@bytedance.com
}

\usepackage{bibentry}

\begin{document}

\twocolumn[{%
\renewcommand\twocolumn[1][]{#1}%
\maketitle
\begin{center}
    \includegraphics[width=0.94\textwidth]{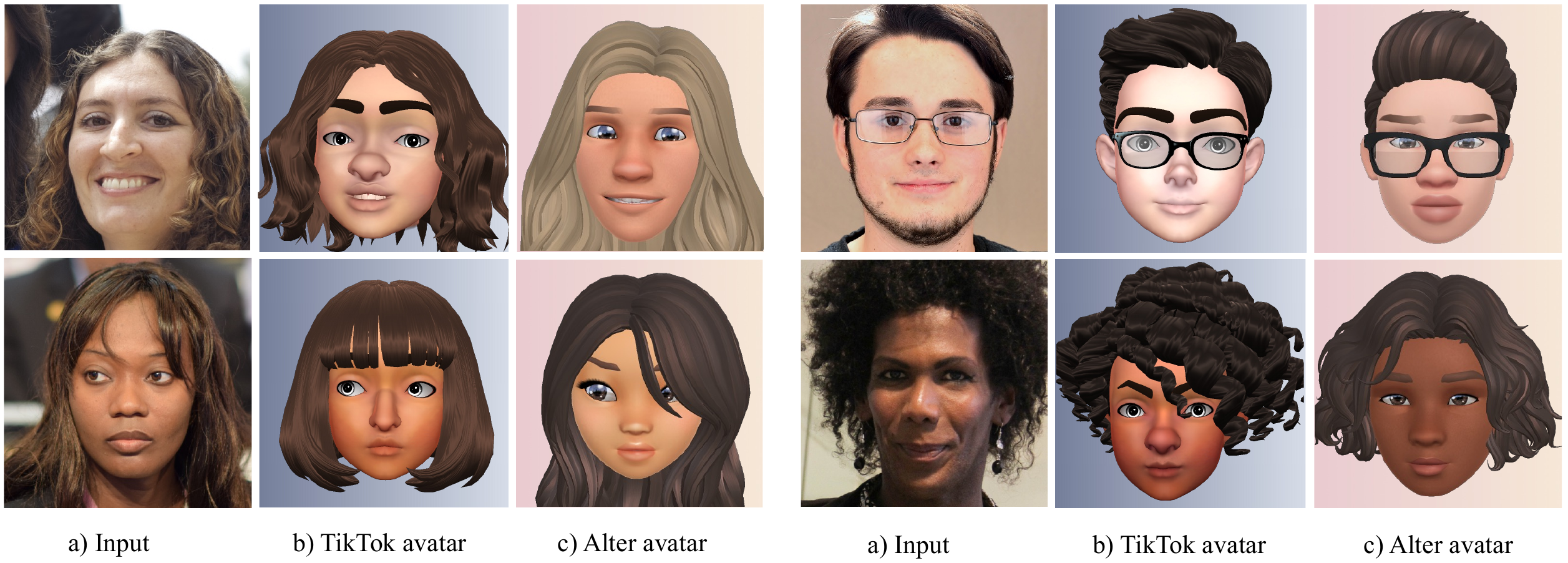}
    \captionof{figure}{a) Given user's front-facing selfie images, SwiftAvatar is able to auto-create corresponding stylized avatars on any arbitrary avatar engine, for example, b) the TikTok avatar engine $^1$, or c) the Alter avatar engine $^2$.}
    \label{fig:teaser}
\end{center}
}]


\begin{abstract}

\let\thefootnote\relax\footnotetext{\hspace{1em} \textsuperscript{\rm *}Equal Contribution}
\let\thefootnote\relax\footnotetext{\hspace{1em} \textsuperscript{\rm $\dagger$}Corresponding Author}
\let\thefootnote\relax\footnotetext{Copyright \copyright \space 2023, Association for the Advancement of Artificial Intelligence (www.aaai.org). All rights reserved.}

The creation of a parameterized stylized character involves careful selection of numerous parameters, also known as the ``avatar vectors" that can be interpreted by the avatar engine. Existing unsupervised avatar vector estimation methods that auto-create avatars for users, however, often fail to work because of the domain gap between realistic faces and stylized avatar images. To this end, we propose \emph{SwiftAvatar}, a novel avatar auto-creation framework that is evidently superior to previous works. SwiftAvatar introduces dual-domain generators to create pairs of realistic faces and avatar images using shared latent codes. The latent codes can then be bridged with the avatar vectors as pairs, by performing GAN inversion on the avatar images rendered from the engine using avatar vectors. 
Through this way, we are able to synthesize paired data in high-quality as many as possible, consisting of avatar vectors and their corresponding realistic faces. We also propose semantic augmentation to improve the diversity of synthesis. 
Finally, a light-weight avatar vector estimator is trained on the synthetic pairs to implement efficient auto-creation. Our experiments demonstrate the effectiveness and efficiency of SwiftAvatar on two different avatar engines. The superiority and advantageous flexibility of SwiftAvatar are also verified in both subjective and objective evaluations.

\end{abstract}

\section{Introduction}

\let\thefootnote\relax\footnotetext{\hspace{1em} \textsuperscript{\rm 1 }https://newsroom.tiktok.com/en-us/express-yourself-through-tiktok-avatars}
\let\thefootnote\relax\footnotetext{\hspace{1em} \textsuperscript{\rm 2 }https://github.com/facemoji/alter-core}

The emerging of the Metaverse concept is alongside with the wide usage of virtual avatars, embodiment of Metaverse users in various styles, which are popular in modern digital lives such as socialization, e-shopping and gaming.
Even though many avatar platforms, such as Zepeto, BitMoji and ReadyPlayerMe, have enabled users to create their own stylized avatars by specifying avatar vectors that can be interpreted by their avatar engines, the manual creation process is tiresome and time consuming, especially when the engine provides with a large set of options making up a long avatar vector. 
Besides, the avatar vectors to specify usually consists of parameters in both continuous forms (to control facial shape, eye spacing, etc.) and discrete forms (to control hair styles, wearings etc.), all requiring careful selection and cautious adjustment to achieve a satisfactory result. 
Therefore, it is valuable to study \emph{how to automatically create a stylized avatar that best matches the user's input selfie}. 

A straightforward way to realize avatar auto-creation is using supervised learning: a network is trained on labeled data to predict avatar vectors based on real face inputs. However, this requires large amount of data collection and manual labeling, which is laborious, expensive, and is not generalizable across engines. Because the definition of avatar vectors and assets vary from engine to engine, data labeled for one engine can not be used to train on other engines. 

\begin{figure*}[t]
\centering
\includegraphics[width=0.98\textwidth]{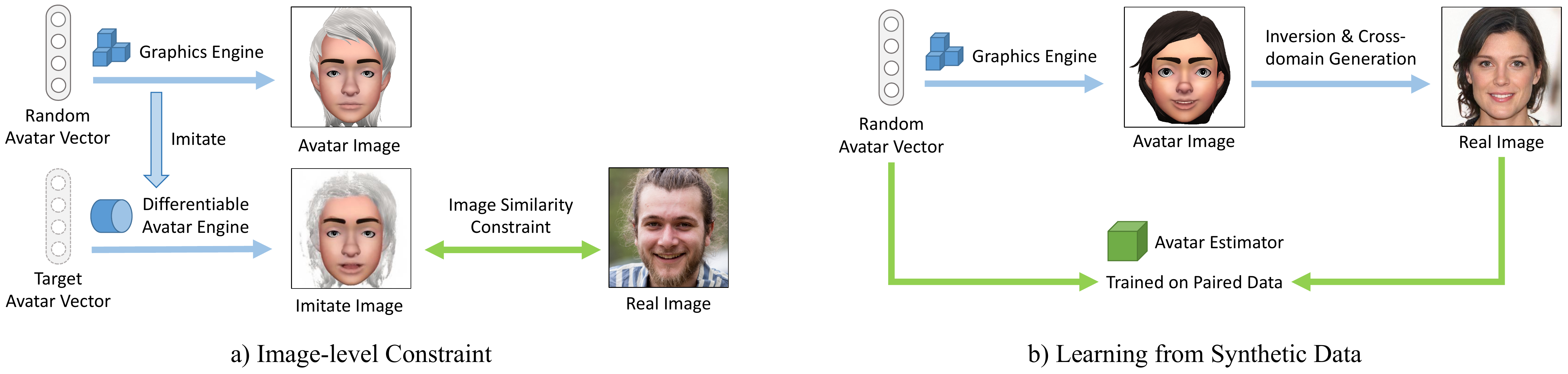}
\caption{Conceptual comparison of previous and our methods. a) Previous methods try to solve unknown avatar vectors under image-level constraints, which requires to train a differentiable engine imitator to participate the optimization. b) While our method can use origin avatar engine and cross-domain generation to synthesize paired data (image-label), and then a model is trained based on the synthetic dataset.}
\label{fig:difference}
\end{figure*}

Several unsupervised learning methods have been proposed to address avatar auto-creation without using any labeled data, including Tied Output Synthesis (TOS) \cite{wolf2017unsupervised} and the Face-to-Parameter (F2P) series \cite{shi2019face, shi2020fast}.
The main idea of these works can be abstracted by Fig. \ref{fig:difference}-a.
In order to achieve an avatar vector that renders avatar image as similar to the input face as possible, these methods impose constraints on the image-level.
They suffer from issues as follows:
1) The image-level similarity constraints they establish are designed for realistic avatar images, not applicable to stylized avatar images that have domain gap with real face images.
2) Leveraging the image-level supervision requires that the avatar rendering process is differentiable. So they usually introduce an imitator network that imitates the behavior of the non-differentiable avatar engine. However, the un-avoidable deviation of imitators from the original avatar engine, as illustrated by Fig. \ref{fig:difference}-a, degrades the accuracy of the similarity measure.
3) Some approaches like F2P \cite{shi2019face} need iterative optimization to guarantee the quality of estimated avatar vectors, which is time-consuming in inference. 

To address these issues, we propose \emph{SwiftAvatar}, a novel avatar auto-creation framework shown by Figure. \ref{fig:difference}-b). Unlike previous works that use similarity constraints to find the avatar vector whose rendered image best matches a given face, the core idea of our framework is cross-domain data synthesis. SwiftAvatar is able to synthesize pairs of avatar vectors and corresponding realistic faces in high fidelity as many as possible. They are used to train a light-weight estimator that directly predicts avatar vectors from input selfie. Specifically, the SwiftAvatar framework consists of three components: \emph{dual-domain generators}, a pipeline for \emph{cross-domain paired data production}, and an \emph{avatar estimator}. 

The dual-domain generators comprise a realistic generator and an avatar generator, both adopting an architecture from SemanticStyleGAN~\cite{shi2022semanticstylegan}. 
The realistic generator is pretrained on real faces, but the avatar generator is transfer-learned on engine-rendered avatar images with color consistency constraints. So that given a same latent code, the two generators could generate a realistic face image and an avatar image that naturally look similar.
Data synthesis relies on the dual-domain generators. 
The production of synthetic data starts from randomly sampled avatar vectors. They are sent to the avatar engine to render avatar images, which are then inverted into latent codes through the avatar generator. Finally, the latent codes are fed into the realistic generator to get realistic face images corresponding to the sampled avatar vectors. 
Moreover, we introduce semantic augmentation to expand the diversity of produced data by adding local perturbation to latent code. 
With the synthetic data, we can then train an avatar estimator on them. 
The avatar estimator is light-weight and efficient at inference. Given user selfie image, it accomplish the auto-creation by directly predicting an avatar vector that matches the input.  

Our experiments involve both objective and subjective evaluations, comparing SwiftAvatar with previous methods in different aspects. 
SwiftAvatar achieves advantageous results over all existing methods in creating avatars fidelity to the input images. Moreover, experiments on two diverse avatar engines verify the strong generality of SwiftAvatar. Qualitative results illustrated in Fig. \ref{fig:teaser}, show that SwiftAvatar can generate reasonable avatars for both engines given input faces. 
In summary, our contributions are as follows:

\begin{itemize}

\item A novel framework, SwiftAvatar, is proposed that can automatically create a stylized avatar given a user selfie image. It can be swiftly applied to any arbitrary avatar engines without extra assumptions (e.g. differentiable rendering capability) on that engine.

\item SwiftAvatar presents a novel pipeline that produces paired data across domains. It involves dual-domain generators to address the domain gap between realistic faces and stylized avatars. A novel semantic augmentation is also devised to improve the diversity of data synthesis.

\item Experiments show that SwiftAvatar outperforms previous methods in terms of both quality and efficiency. Results on two different avatar engines also verify the strong generalizability of SwiftAvatar. 

\end{itemize}

\section{Related Work}

\subsubsection{3D Face Reconstruction}

Many progresses on 3D face reconstruction \cite{tuan2017regressing,dou2017end} cannot be achieved without 3D morphable models (3DMM) \cite{blanz1999morphable} and its variants like BFM \cite{gerig2018morphable} and FLAME \cite{FLAME:SiggraphAsia2017}, where the geometry and texture of a photorealistic 3D face are parameterized as a vector through linear transform.
Recent researches also explore representing 3D faces in other formats like dense landmarks \cite{feng2018evaluation} or position maps \cite{feng2018joint}.
However, stylized avatar auto-creation is not 3D face reconstruction. They differ in two folds: 1) 3D face reconstruction methods aim to recover realistic 3D faces, not stylized characters;
2) Most avatar engines utilize avatar vectors to render stylized 3D characters according to manually designed assets. Estimating avatar vectors is much more difficult than reconstructing 3DMM coefficients, since the graphics rendering of avatar engines are usually black-box and non-differentiable.

\subsubsection{GAN and GAN Inversion}

The rapid growth of generative networks, such as GANs \cite{goodfellow2014generative,karras2017progressive} and VAEs \cite{kingma2013auto,razavi2019generating}, inspires methods to use \emph{latent codes} that can implicitly parameterize face images through a pre-trained generator. 
Among the pre-trained generators, the most popular one is the generator of the StyleGAN \cite{karras2019style,karras2020analyzing}, known for its impressive capability in generating high quality face images.
In order to leverage the pre-trained StyleGAN generator in editing existing face images, GAN inversion \cite{zhu2020domain} is required to compute the latent code whose StyleGAN output best matches a given face image. Existing GAN inversion methods estimate latent codes using either iterative optimization \cite{abdal2020image2stylegan++} or training a feed-forward encoder \cite{richardson2021encoding} using VGG perceptual loss \cite{johnson2016perceptual}, LPIPS loss \cite{zhang2018unreasonable} or face identity loss \cite{deng2019arcface}. 
Though GAN inversion can not be directly applied to estimate the avatar vectors due to the hand-crafted nature of avatar engines, SwiftAvatar leverages GAN inversion in its data production pipleline to build the correspondence between avatar vectors and latent codes.

\subsubsection{Portrait Stylization}

Recent literatures on portrait stylization also benefit a lot from pre-trained StyleGANs.
On one hand, finetuning the pre-trained StyleGAN generator provides with an agile approach for synthesizing faces of new styles \cite{song2021agilegan,back2021fine}.
On the other hand, freezing low-level layers of the generator during finetuning helps preserving structural information (e.g. face attributes) between face images generated from the original generator (realistic domain) and the finetuned generator (stylized domain), when they are using similar latent codes \cite{back2021fine,huang2021unsupervised}.
Though this {cross-domain latent code sharing strategy} cannot be directly applied on avatar engines, we discover that it has a strong potential in cross-domain data synthesis. 
In specific, we design the data production pipeline of SwiftAvatar to leverage a pre-trained SemanticStyleGAN \cite{shi2022semanticstylegan}. It adopts a compositional generator architecture, disentangling the latent space into different semantic areas separately, and could provide more precise local controls on synthesized face images.

\subsubsection{Stylized Avatar Auto-Creation}

Auto-creating avatars for individual users has become an important capability to the entertainment industry.To ensure quality, commercial solution usually involve a large amount of manual annotations, something this paper seeks to avoid.
Among the published approaches that avoid manual annotations,
some \cite{shi2019face,shi2020fast} are designed for realistic avatars only, which share the same domain with the real face images, it is easy to define whether they match or not.
Therefore, these methods can utilize straightforward in-domain supervisions to improve fidelity of creation, such as the image-level L1/L2 losses, face parsing loss, or the face identity loss \cite{deng2019arcface}.
Creating stylized avatars from real faces, on the contrary, is more difficult than creating realistic avatars. 
There are only few works toward this direction.
Tied Output Synthesis (TOS) \cite{wolf2017unsupervised} devises a encoder-decoder structure, that eliminates the domain gap by sharing one encoder across two domains. 
The contemporaneous work AgileAvatar \cite{sang2022agileavatar} formulates a cascaded framework which progressively bridges the domain gap.
To address the non-differentiable issue of avatar engines, all these methods need to train an imitator network to imitate the engines' rendering procedure.
While the quality of the images generated by the imitator is relatively poor. Besides, they all impose explicit constraints on images from different domains, resulting in suboptimal matches because of the domain gap.
By contrast, our SwiftAvatar framework employs original avatar engine to generate high-quality images and utilizes dual-domain generators to overcome the domain gap.

\begin{figure*}[h]
\centering
\includegraphics[width=0.98\textwidth]{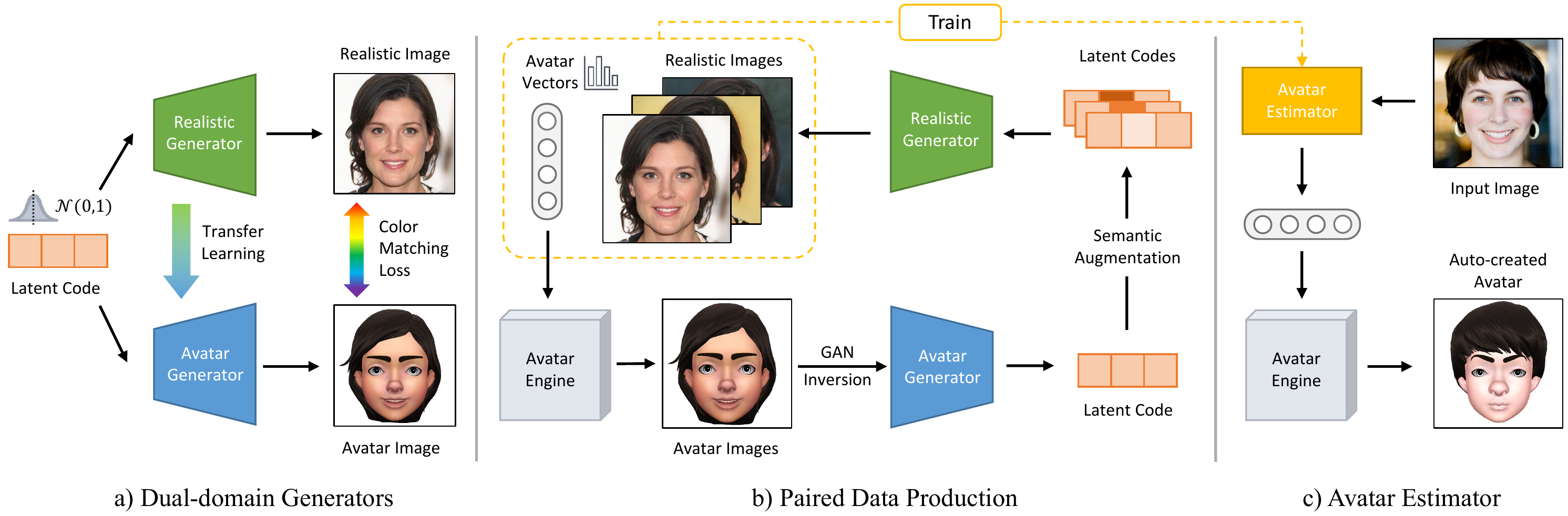}
\caption{Overview of our method. 
a) Dual-domain generators consist of a fixed realistic generator and a transfer-learned avatar generator. They can generate corresponding realistic and avatar images when given the same latent codes. 
b) Paired data production starts from randomly sampled avatar vectors, which are sent to the avatar engine to render avatar images. These images are then inverted into latent codes through
GAN inversion, and the latent codes are fed into the realistic generator to get realistic face images corresponding to the sampled avatar vectors.
c) Supervised by the paired data, the avatar estimator can predict avatar vectors when given an real face input, then the avatar can be rendered by avatar engine.}
\label{fig:method}
\end{figure*}

\section{Methodology}

In this section, we present our unsupervised framework for stylized avatar's auto-creation, as shown in Fig.\ref{fig:method}.
It aims at estimating avatar vector $p$ for input real face image $x$, then the estimated $p$ can be used  to render the corresponding stylized avatar $y$ by avatar engine $E$. Our solution is split into three parts: dual-domain generators in Sec. \ref{sec:generator}, paired data production in Sec. \ref{sec:production}, and avatar estimator in Sec. \ref{sec:estimator}.
Dual-domain generators address the domain gap problem by generating realistic faces and stylized avatars with shared latent codes. Then, paired data production focuses on how to generate paired data consisting of avatar vectors and realistic faces. Finally, avatar estimator estimates desired avatar vectors similar to input real face images. 

\subsection{Dual-domain Generators}
\label{sec:generator}

The dual-domain generators consist of a realistic generator $G_\textit{real}$ and an avatar generator $G_\textit{avatar}$ to achieve cross-domain image generation. Given the same latent code, they can simultaneously synthesize paired images of both realistic face and stylized avatar while preserving the same attributes (e.g. skin color, hair style). To impose an extra facial consistency constraint between the realistic and the avatar domains, we adopt SemanticStyleGAN\cite{shi2022semanticstylegan} as the architecture of two generators owing to its extra semantic segmentation output.

\subsubsection{Cross-domain Generation}


Pretrained SemanticStyleGAN on CelebAMask-HQ \cite{CelebAMask-HQ} is directly used as realistic generator $G_\textit{real}$, and also used to initialize the weight of $G_\textit{avatar}$.
We perform transfer learning on $G_\textit{avatar}$: using only limited number of avatar images $\mathcal{Y}$ to finetune avatar generator $G_\textit{avatar}$. 
$\mathcal{Y}$ are rendered from avatar engine using randomly sampled avatar vectors. 
The finetune procedure follows the settings in SemanticStyleGAN, which using the loss of StyleGAN2 \cite{karras2020analyzing}:
\begin{equation}
    \mathcal{L}_{StyleGAN2} = \mathcal{L}_{adv} + \lambda_{R1} \mathcal{L}_{R1} + \lambda_{path} \mathcal{L}_{path}
\end{equation}
where $\lambda_{R1}$, $\lambda_{path}$ are the constant weights of R1 regularization \cite{mescheder2018training} and path length regularization \cite{karras2020analyzing} separately.
Adversarial loss $\mathcal{L}_{adv}$ adopts non-saturating logistic loss \cite{goodfellow2014generative} and it forces $G_\textit{avatar}$ to generate images similar to avatar image dataset $\mathcal{Y}$.
R1 regularization $\mathcal{L}_{R1}$ is employed to improve the training stability and reduce the number of artifacts.
And path length regularization $\mathcal{L}_{path}$ leads to more reliable and consistently behaving models.

\subsubsection{Facial Consistency}
Although directly fine-tuning SemanticStyleGAN can generate structurally similar paired data of avatar image and realistic face, the colors of each region are not well matched. 
Since SemanticStyleGAN learns the joint modeling of image and semantic segmentation. Such design can simultaneously synthesize face images and their semantic segmentation results. We utilize the semantic segmentation output and introduce a color matching loss for cross-domain facial color consistency.
Specifically, we extract specified pixels from same semantic areas in generated paired images, and match the mean color of them. $m^s(I)$ is the mean color of the region $s$ in the image $I$, and we mainly consider matching the color in hair and skin areas: $s\in \{ hair, skin \}$. The color matching loss is:
\begin{equation}
\mathcal{L}_{color} = \sum_s \left \| m^s(G_\textit{real}(z)) - m^s(G_\textit{avatar}(z)) \right \| ^2
\end{equation}

Overall, the final finetuning loss $\mathcal{L}_{total}$ for $G_\textit{avatar}$ is formulated as:
\begin{equation}
\mathcal{L}_{total} = \mathcal{L}_{StyleGAN2} + \lambda_{color} \mathcal{L}_{color}
\end{equation}

\begin{figure}[t]
\centering
\includegraphics[width=0.46\textwidth]{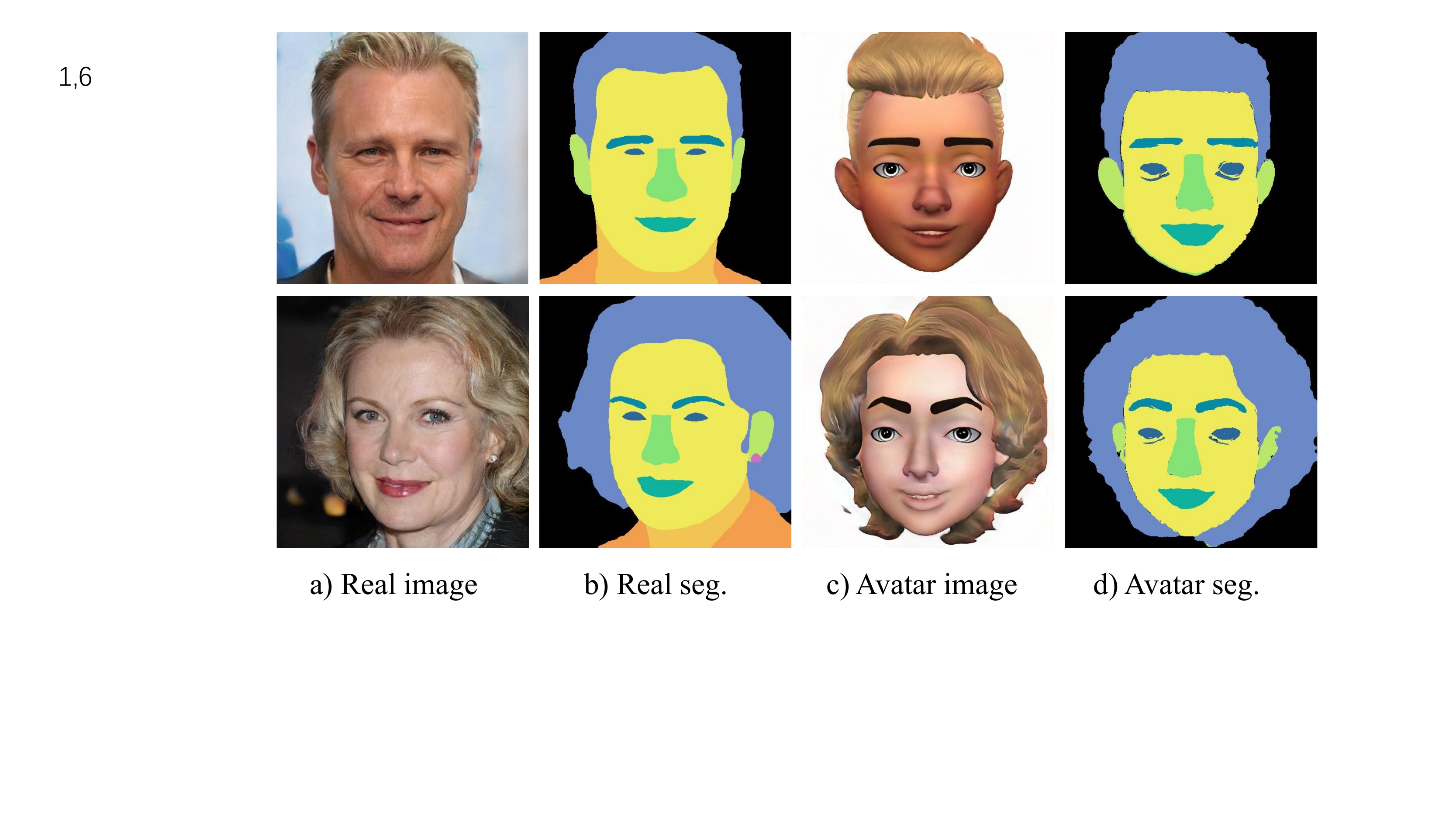}
\caption{Generated results of our dual-domain generators. When providing randomly sampled latent codes, we can obtain a) and b), realistic faces and their semantic segmentation results. Simultaneously, we can obtain c) and d), stylized avatar images and their semantic segmentation results.}
\label{fig:finetune}
\end{figure}

An example is shown in Figure. \ref{fig:finetune}, when given a shared latent code, the dual-domain generators can generate a pair of images: a realistic face and a stylized avatar. They are similar in facial structure and color composition, yet belong to two different domains.

\subsection{Paired Data Production}
\label{sec:production}

Paired data production pipeline focuses on synthesizing paired avatar vectors and realistic face images, as is illustrated in Figure. \ref{fig:method}. We sample a number of random avatar vectors as labels $\mathcal{P}$, which are used by the graphics engine to generate corresponding avatar images $\mathcal{Y}$. Then, for every avatar image $y$, we pass it through GAN inversion to get its latent code $w$. We adopt optimization-based GAN inversion for its better performance:
\begin{equation}
w^* = \mathop{\arg\min}\limits_{w} \mathcal{L}_{invert}(y, w)
\end{equation}

where $w^*$ represents our target latent code. For faster convergence, latent codes are initialized with mean value $w_{mean}$, and optimized by gradient descent. We use LPIPS loss  \cite{zhang2018unreasonable} to measure perceptual similarity, and mean squared error (MSE) loss between original images and reconstructed avatar images to measure reconstruction similarity. Besides, MSE loss between $w_{mean}$ and $w$ is set as latent regularization to ensure better generation results. The loss function is formulated as:
\begin{align}
\mathcal{L}_{invert} = \ & \lambda_p \ \textit{LPIPS}( y,  G_\textit{avatar}(w) ) + \nonumber \\
& \lambda_i \left \| y - G_\textit{avatar}(w) \right \| ^2 + \lambda_l \left \| w - w_{mean} \right \| ^2
\end{align}
where $\lambda_p$, $\lambda_i$, $\lambda_l$ are constants to balance three loss terms. After obtaining desired latent code $w$, it is fed into the realistic generator $G_\textit{real}$ to generate a realistic face $x$ which is similar to original avatar image $y$ in identification: 

\begin{equation}
x = G_\textit{real}(w)
\end{equation}

In this way, we can generate paired data $(p,x)$ used for later avatar vector estimation training process. An example illustration of paired data is shown in Figure. \ref{fig:invert}. 

\subsubsection{Semantic Augmentation}

\begin{figure}
\centering
\includegraphics[width=0.46\textwidth]{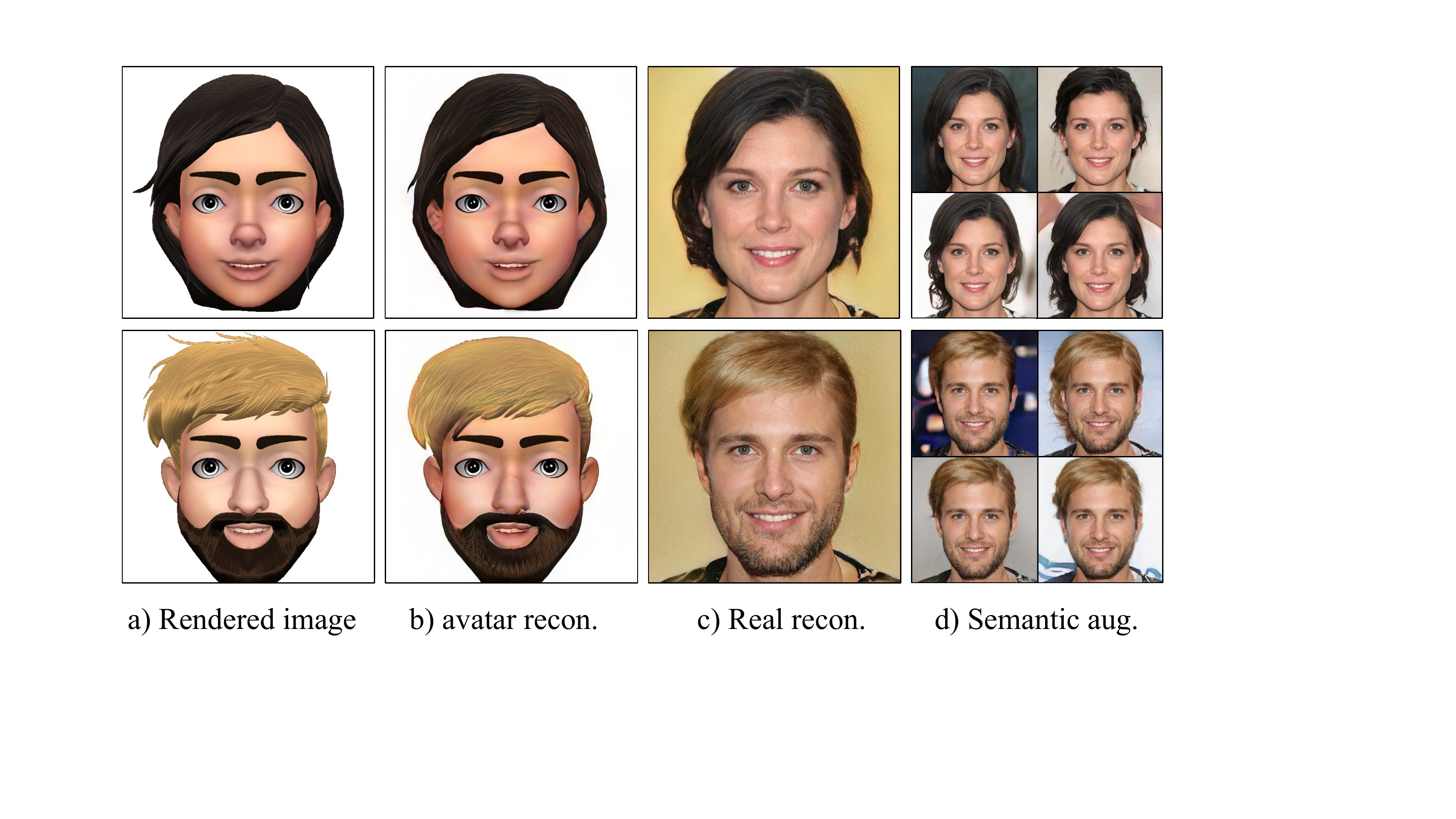}
\caption{Examples of produced paired data. a) Engine rendered avatar images. b) Reconstructed avatar images. c) Generated realistic images. d) Semantic augmented images.}
\label{fig:invert}
\end{figure}

The sampled avatar vectors, as well as their rendered avatar faces suffer from the lack of diversity, due to the limited amount of assets available for avatar engines.
Take the ``hair style'' attribute for example, in most stylized avatar engines, different hair styles are determined by selecting from different hair meshes all predefined in the engine. The limited number of predefined mesh assets restricts the capability of avatar engines to match a real-world user face, whose hair styles would be countless. To enrich the diversity of generated realistic faces in data production, we take advantage of the compositional generation ability of our dual-domain generators which are implemented as SemanticStyleGANs, and design the \emph{semantic augmentation}. In SemanticStyleGAN, each semantic part is modulated individually with corresponding local latent codes. Such property enables us to manipulate only specific regions of the synthetic images while keeping other regions unchanged. In implementation, we add random noises to part of latent code corresponding to these ambiguous avatar vectors (e.g. hair type). Semantic augmentation can be described as: 
\begin{equation}
w_k \to (1-\lambda_{aug}) w_k + \lambda_{aug} n
\end{equation}
where $\lambda_{aug}$ is a hyper-parameter to adjust semantic augmentation tensity, $w_k$ is the local part of latent code to be augmented, and $n$ represents for random noise. A semantic augmentation example is shown in Figure. \ref{fig:invert}, where hair and background region are changed.

\subsection{Avatar Estimator}
\label{sec:estimator}

Once the aforementioned synthetic paired dataset is produced, we can train an avatar estimator to predict avatar vectors, which contain continuous and discrete parameters. We choose ResNet-18 (He et al. 2016) pretrained on ImageNet (Deng et al. 2009) as our backbone. We remove its last fully connected layers and add multiple separate MLP heads for different parameter estimation. 
All continuous parameters form a target vector to be predicted in one head, supervised by L1 loss. 
Every discrete parameter estimation is carried out with a standalone head. 
Because both generation and semantic augmentation would inevitably introduce noises to discrete labels, we choose symmetric cross-entropy (SCE) loss \cite{wang2019symmetric}, which has been proven robust to noises, for the optimization of discrete tasks. 
The total loss of avatar estimator is:
\begin{equation}
\mathcal{L}_{estimator} = \lambda_d \ \textit{SCE}(\widehat{p_d^i}, p_d^i) + \lambda_c \left | \widehat{p_c} - p_c \right | 
\end{equation}
where $\lambda_d$ and $\lambda_c$ are hyper-parameters to balance two loss terms. $\widehat{p_d^i}$, $\widehat{p_c}$ are the prediction results of $i$-th discrete head and continuous head respectively. And ${p_d^i}$, ${p_c}$ are their corresponding ground-truth.

\begin{figure*}[th]
\centering
\includegraphics[width=0.915\textwidth]{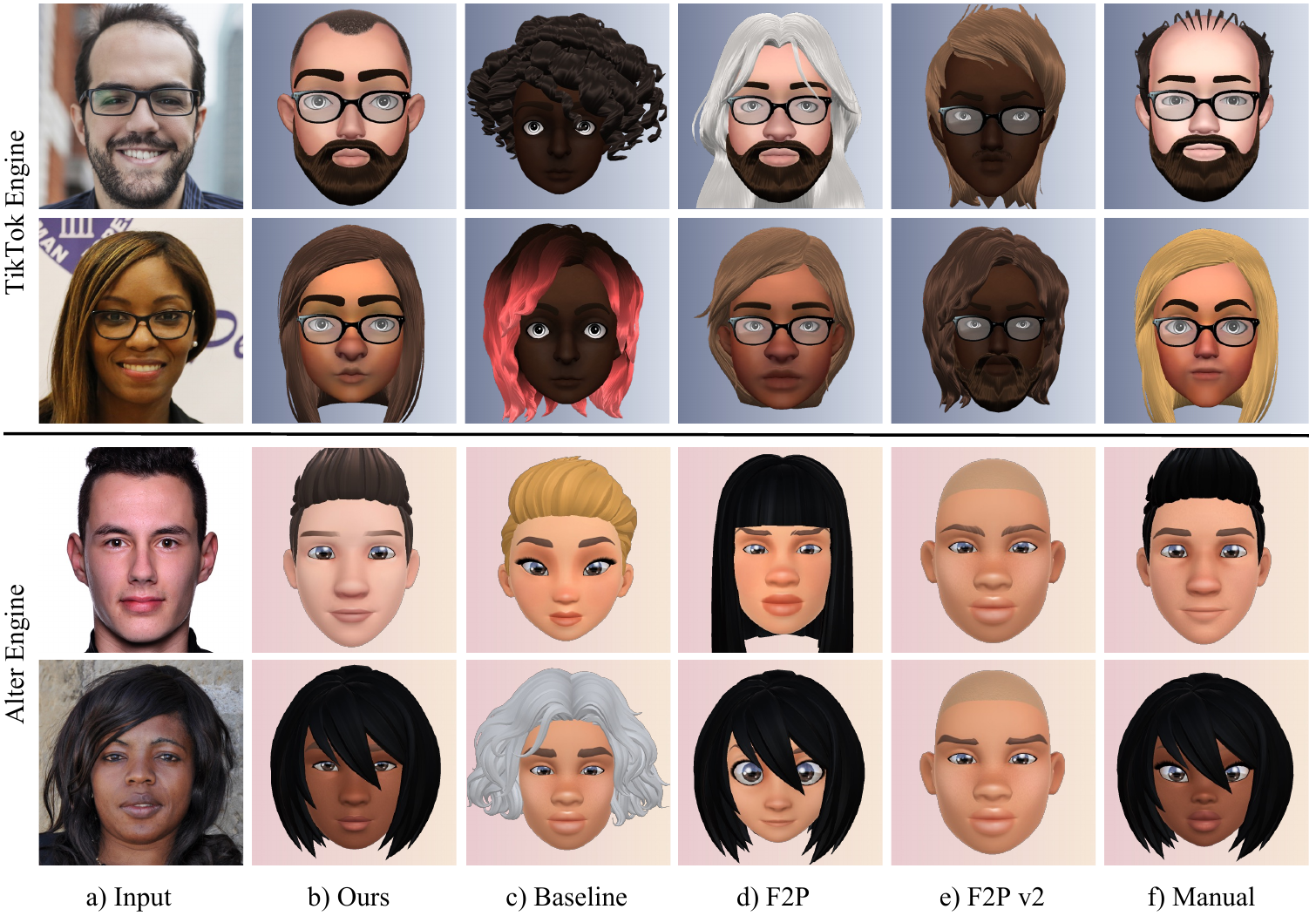}
\caption{Visual comparison with other methods on TikTok engine and Alter engine. a) Given input images, b) our method creates stylized avatars that look similar to input images. c) Baseline method without considering the domain gap problem in stylized avatars. d) F2P \cite{shi2019face}, an image-level constraint method intending to create semi-realistic avatars, is not suitable to create stylized avatars. e) F2P v2 \cite{shi2020fast}, a fast and robust version of F2P. f) Manually created avatars by professional designers, which can be regarded as ground truth.}
\label{fig:comparison}
\end{figure*}

\section{Experiments}

\subsubsection{Experimental Data}
To verify the effectiveness of our method, we conduct experiments on two stylized avatar engines: the TikTok engine and the Alter engine. The TikTok engine contains resourceful discrete and continuous avatar parameters, so we generate 50000 avatar vectors and corresponding rendered images. The Alter engine is an open-source avatar engine and only contains discrete avatar parameters which has fewer assets than the TikTok engine, so we just generate 10000 avatar vectors and corresponding rendered images. The detailed information of both avatar engines can be found in supplementary materials. These avatar images and avatar vectors are used for finetuning avatar generator and producing paired data. For evaluation, we choose 116 images from FFHQ dataset \cite{karras2019style}, which consists of diverse kinds of face shapes, hairstyles, etc. We invite designers to manually generate avatar vectors for these 116 images as ground truth.

\subsubsection{Implementation Details}
We implement our methods using PyTorch 1.10 library and perform all experiments on NVIDIA V100 GPUs. 
When finetuning the avatar generator $G_\textit{avatar}$, we use the same optimizer settings as in SemanticStyleGAN. Batch size is set to 16, style mixing probability \cite{karras2019style} is set to 0.3.
$\lambda_{R1}$, $\lambda_{path}$ are set to 10 and 0.5 separately.
Lazy regularization \cite{karras2020analyzing} is applied every 16 mini-batches for discriminator (R1 regularization) and every 4 mini-batches for generator (path length regularization). All the images used for generators are aligned and resized to resolution $512 \times 512$.
The optimization-based GAN inversion approach employs Adam \cite{kingma2014adam} optimizer in the paired data production stage, and the learning rate initially follows cosine annealing with 0.1.
We optimize 200 steps for all latent codes, and $\lambda_i$, $\lambda_p$, $\lambda_l$ are set to 0.1, 1 and 1, respectively. The mean latent code $w_{mean}$ is the average among $10^4$ randomly generated latent codes from avatar generator $G_\textit{avatar}$ in $\mathcal{W}$ space, and serve as the initialization. Notably, directly optimizing the latent code could be problematic since some avatar assets are transparent, e.g. glasses. Thus, we use a modified version $\tilde{w}_{mean}$ for latent code initialization (See supplementary materials for details).
For semantic augmentation, we generate 10 augmented images for each latent code, using randomly generated noise in $\mathcal{W}$ space.we set $\lambda_{aug}$ to 1 for the background to improve the model robustness of background variance, and also set $\lambda_{aug}$ to 0.3, 0.06 for the hair part and glasses part to expand data diversity.
In the avatar estimator training stage, the input images of avatar estimator are resized to $224 \times 224$. We use the Adam optimizer with batch size 256 to train 100 epochs. The learning rate is set to $1e-3$, and decayed by half per 30 epochs. For the experiments on TikTok engine, there are 1 continuous head and 8 discrete heads inside the avatar estimator, so we set $\lambda_c$ and $\lambda_d$ to 1 and 10 separately. For the experiments on Alter engine, the avatar estimator contains 6 discrete heads and the training loss only contains discrete loss.

\subsection{Comparison with Other methods}

We compare our method with other methods, including Baseline, F2P \cite{shi2019face}, F2P v2 \cite{shi2020fast} on both their auto-creation similarity and inference speed. Baseline method is setup as ignoring the domain gap problem, where the avatar estimator is trained on $(y,p)$ paired data, that is, trained on rendered avatar images instead of real face images. Figure. \ref{fig:comparison} shows a comparison of rendered stylized avatars among different methods corresponding to their predicted avatar vectors. As can be seen, our method can better address the domain gap problem and create stylized avatars with high similarity with input real faces and approximate the quality of manual method from designers. For more results, please refer to supplementary materials.

\begin{table}
\centering

\begin{tabular}{l|l|l|l}
    \toprule
    Method     & TikTok engine $\downarrow$    & Alter engine $\downarrow$      & Speed $\uparrow$  \\
    \midrule
    Baseline   & 0.5033            & 0.3962            & $\sim 10^2$ Hz     \\
    F2P        & 0.3562            & 0.3040            & $\sim 1$ Hz    \\
    F2P v2     & 0.4466            & 0.2825            & $\sim 10^2$ Hz    \\
    Ours       & \textbf{0.3110}   & \textbf{0.2405}   & \textbf{$\mathbf{\sim 10^2}$ Hz}  \\
    \bottomrule
\end{tabular}
\caption{Quantitative Evaluation. We compare our method with three other methods on TikTok engine and Alter engine. Lower distance represents results are more similar to manual-creation. We test speed on NVIDIA V100.}
\label{tab:evaluation}
\end{table}

\subsection{Quantitative Evaluation}
\label{sec:evaluation}
Although the avatar auto-creation has no standard answer, we generally regard the avatar manually created by professional designers as the ground truth, and quantitatively evaluate at image level. We calculate the perceptual distance (LPIPS) \cite{huang2021unsupervised} between auto-created avatar and manual-created avatar to simulate human observation. The lower distance indicates the avatar is better matched to input image. The results are presented in Table. \ref{tab:evaluation}, from which we can see our method significantly outperforms others. Since our method only needs one forward propagation, our inference speed is also competitive, which can be applied in real-time applications.

\subsection{Human Rating}
We invited 50 volunteers to subjectively evaluate all algorithms of 50 images from evaluation dataset on both graphics engine. Given the input real face images and stylized avatars generated from our and other methods, the volunteers were requested to pick the best matching avatar for each input real face. The methods were presented in random order, and the volunteers were given an unlimited time to choose. The human rating results are shown in Table. \ref{tab:human}. $59.11\%$ of the answers on TikTok engine and $63.68\%$ on the Alter engine selected our method as the best matching results, showing the superiority of our method compared with others.

\begin{table}
\centering
\begin{tabular}{l|l|l|l|l}
    \toprule
    Method      & Ours     & baseline & F2P    & F2P v2  \\
    \midrule
    TikTok      & \textbf{59.11}\%      & 5.66\%      & 30.86\%    & 4.37\%     \\
    Alter       & \textbf{63.68}\%      & 18.94\%      & 8.65\%    & 8.73\%     \\
    \bottomrule
\end{tabular}
\caption{Human subjective rating on two engines. Our method earns the most choices when asked to choose the avatar which matches the human image.}
\label{tab:human}
\end{table}

\begin{table}[t]
\centering
\begin{tabular}{l|l|l}
    \toprule
    Method               & TikTok engine $\downarrow$  & Alter engine $\downarrow$ \\
    \midrule
    baseline             & 0.5033         & 0.3962 \\
    + domain adaptation  & 0.3401         & 0.3123 \\
    + semantic aug       & \textbf{0.3110}         & \textbf{0.2405}  \\
    \bottomrule
\end{tabular}
\caption{Ablation study on two engines. Lower distance indicates better matching results to manual-creation.}
\label{tab:ablation}
\end{table}

\subsection{Ablation Study}
We conduct the ablation experiments in terms of domain adaptation and semantic augmentation to verify their importance in our framework. We adopt the same evaluation metric described in Sec. \ref{sec:evaluation}. The ablation starts from Baseline method, where the avatar estimator is trained on rendered avatar images and avatar vectors pair. Then on the basis of baseline method, we add domain adaptation and semantic augmentation in turn. Table. \ref{tab:ablation} shows their quantitative results on different engines. The domain adaptation greatly alleviates the domain gap problem in stylized avatar auto-creation, establishes a bridge of connecting real faces and stylized avatars. The semantic augmentation brings noticeable improvement for our method due to its expansion of diversity to samples.

\section{Limitations and Future Work}
There are two main limitations we observed in the experiments. First, our method occasionally predicts wrong color influenced by environmental lighting. This problem might be resolved by considering lighting condition into our pipeline. Second, in paired data production, avatar vector sampling distribution directly influences the training data quality. Simply random sampling could produce some strange images and may cause long-tail problem (e.g. gender, age). In the future, we will perform attribute analysis, and introduce reasonable sampling priors to address synthetic data distribution problem.

\section{Conclusion}
In summary, we present a novel unsupervised framework for auto-creation of stylized avatars. We design the dual-domain generators to address the domain gap between the real images and stylized avatars. Then following the paired data production pipeline, high-quality paired data are produced, which is used for training the avatar estimator. Finally stylized avatars are created by conducting efficient avatar vector estimation. Compared with previous methods, our method is more concise in training stage and more efficient in inference stage. Results on quantitative evaluation and human rating demonstrate the superiority of our method. Also, the success of applying on two different avatar graphics engines demonstrates the generality of our method.

\bibliography{aaai23}

\begin{thebibliography}{31}
\providecommand{\natexlab}[1]{#1}

\bibitem[{Abdal, Qin, and Wonka(2020)}]{abdal2020image2stylegan++}
Abdal, R.; Qin, Y.; and Wonka, P. 2020.
\newblock Image2stylegan++: How to edit the embedded images?
\newblock In \emph{Proceedings of the IEEE/CVF conference on computer vision
  and pattern recognition}, 8296--8305.

\bibitem[{Back(2021)}]{back2021fine}
Back, J. 2021.
\newblock Fine-Tuning StyleGAN2 For Cartoon Face Generation.
\newblock \emph{arXiv preprint arXiv:2106.12445}.

\bibitem[{Blanz and Vetter(1999)}]{blanz1999morphable}
Blanz, V.; and Vetter, T. 1999.
\newblock A morphable model for the synthesis of 3D faces.
\newblock In \emph{Proceedings of the 26th annual conference on Computer
  graphics and interactive techniques}, 187--194.

\bibitem[{Deng et~al.(2019)Deng, Guo, Xue, and Zafeiriou}]{deng2019arcface}
Deng, J.; Guo, J.; Xue, N.; and Zafeiriou, S. 2019.
\newblock Arcface: Additive angular margin loss for deep face recognition.
\newblock In \emph{Proceedings of the IEEE/CVF conference on computer vision
  and pattern recognition}, 4690--4699.

\bibitem[{Dou, Shah, and Kakadiaris(2017)}]{dou2017end}
Dou, P.; Shah, S.~K.; and Kakadiaris, I.~A. 2017.
\newblock End-to-end 3D face reconstruction with deep neural networks.
\newblock In \emph{proceedings of the IEEE conference on computer vision and
  pattern recognition}, 5908--5917.

\bibitem[{Feng et~al.(2018{\natexlab{a}})Feng, Wu, Shao, Wang, and
  Zhou}]{feng2018joint}
Feng, Y.; Wu, F.; Shao, X.; Wang, Y.; and Zhou, X. 2018{\natexlab{a}}.
\newblock Joint 3d face reconstruction and dense alignment with position map
  regression network.
\newblock In \emph{Proceedings of the European conference on computer vision
  (ECCV)}, 534--551.

\bibitem[{Feng et~al.(2018{\natexlab{b}})Feng, Huber, Kittler, Hancock, Wu,
  Zhao, Koppen, and R{\"a}tsch}]{feng2018evaluation}
Feng, Z.-H.; Huber, P.; Kittler, J.; Hancock, P.; Wu, X.-J.; Zhao, Q.; Koppen,
  P.; and R{\"a}tsch, M. 2018{\natexlab{b}}.
\newblock Evaluation of dense 3D reconstruction from 2D face images in the
  wild.
\newblock In \emph{2018 13th IEEE International Conference on Automatic Face \&
  Gesture Recognition (FG 2018)}, 780--786. IEEE.

\bibitem[{Gerig et~al.(2018)Gerig, Morel-Forster, Blumer, Egger, Luthi,
  Sch{\"o}nborn, and Vetter}]{gerig2018morphable}
Gerig, T.; Morel-Forster, A.; Blumer, C.; Egger, B.; Luthi, M.; Sch{\"o}nborn,
  S.; and Vetter, T. 2018.
\newblock Morphable face models-an open framework.
\newblock In \emph{2018 13th IEEE International Conference on Automatic Face \&
  Gesture Recognition (FG 2018)}, 75--82. IEEE.

\bibitem[{Goodfellow et~al.(2014)Goodfellow, Pouget-Abadie, Mirza, Xu,
  Warde-Farley, Ozair, Courville, and Bengio}]{goodfellow2014generative}
Goodfellow, I.; Pouget-Abadie, J.; Mirza, M.; Xu, B.; Warde-Farley, D.; Ozair,
  S.; Courville, A.; and Bengio, Y. 2014.
\newblock Generative adversarial nets.
\newblock \emph{Advances in neural information processing systems}, 27.

\bibitem[{Huang, Liao, and Kwong(2021)}]{huang2021unsupervised}
Huang, J.; Liao, J.; and Kwong, S. 2021.
\newblock Unsupervised image-to-image translation via pre-trained stylegan2
  network.
\newblock \emph{IEEE Transactions on Multimedia}, 24: 1435--1448.

\bibitem[{Johnson, Alahi, and Fei-Fei(2016)}]{johnson2016perceptual}
Johnson, J.; Alahi, A.; and Fei-Fei, L. 2016.
\newblock Perceptual Losses for Real-Time Style Transfer and Super-Resolution.
\newblock In Leibe, B.; Matas, J.; Sebe, N.; and Welling, M., eds.,
  \emph{Computer Vision -- ECCV 2016}, 694--711. Cham: Springer International
  Publishing.
\newblock ISBN 978-3-319-46475-6.

\bibitem[{Karras et~al.(2017)Karras, Aila, Laine, and
  Lehtinen}]{karras2017progressive}
Karras, T.; Aila, T.; Laine, S.; and Lehtinen, J. 2017.
\newblock Progressive growing of gans for improved quality, stability, and
  variation.
\newblock \emph{arXiv preprint arXiv:1710.10196}.

\bibitem[{Karras, Laine, and Aila(2019)}]{karras2019style}
Karras, T.; Laine, S.; and Aila, T. 2019.
\newblock A style-based generator architecture for generative adversarial
  networks.
\newblock In \emph{Proceedings of the IEEE/CVF conference on computer vision
  and pattern recognition}, 4401--4410.

\bibitem[{Karras et~al.(2020)Karras, Laine, Aittala, Hellsten, Lehtinen, and
  Aila}]{karras2020analyzing}
Karras, T.; Laine, S.; Aittala, M.; Hellsten, J.; Lehtinen, J.; and Aila, T.
  2020.
\newblock Analyzing and improving the image quality of stylegan.
\newblock In \emph{Proceedings of the IEEE/CVF conference on computer vision
  and pattern recognition}, 8110--8119.

\bibitem[{Kingma and Ba(2014)}]{kingma2014adam}
Kingma, D.~P.; and Ba, J. 2014.
\newblock Adam: A method for stochastic optimization.
\newblock \emph{arXiv preprint arXiv:1412.6980}.

\bibitem[{Kingma and Welling(2013)}]{kingma2013auto}
Kingma, D.~P.; and Welling, M. 2013.
\newblock Auto-encoding variational bayes.
\newblock \emph{arXiv preprint arXiv:1312.6114}.

\bibitem[{Lee et~al.(2020)Lee, Liu, Wu, and Luo}]{CelebAMask-HQ}
Lee, C.-H.; Liu, Z.; Wu, L.; and Luo, P. 2020.
\newblock MaskGAN: Towards Diverse and Interactive Facial Image Manipulation.
\newblock In \emph{IEEE Conference on Computer Vision and Pattern Recognition
  (CVPR)}.

\bibitem[{Li et~al.(2017)Li, Bolkart, Black, Li, and
  Romero}]{FLAME:SiggraphAsia2017}
Li, T.; Bolkart, T.; Black, M.~J.; Li, H.; and Romero, J. 2017.
\newblock Learning a model of facial shape and expression from {4D} scans.
\newblock \emph{ACM Transactions on Graphics, (Proc. SIGGRAPH Asia)}, 36(6):
  194:1--194:17.

\bibitem[{Mescheder, Geiger, and Nowozin(2018)}]{mescheder2018training}
Mescheder, L.; Geiger, A.; and Nowozin, S. 2018.
\newblock Which training methods for GANs do actually converge?
\newblock In \emph{International conference on machine learning}, 3481--3490.
  PMLR.

\bibitem[{Razavi, Van~den Oord, and Vinyals(2019)}]{razavi2019generating}
Razavi, A.; Van~den Oord, A.; and Vinyals, O. 2019.
\newblock Generating diverse high-fidelity images with vq-vae-2.
\newblock \emph{Advances in neural information processing systems}, 32.

\bibitem[{Richardson et~al.(2021)Richardson, Alaluf, Patashnik, Nitzan, Azar,
  Shapiro, and Cohen-Or}]{richardson2021encoding}
Richardson, E.; Alaluf, Y.; Patashnik, O.; Nitzan, Y.; Azar, Y.; Shapiro, S.;
  and Cohen-Or, D. 2021.
\newblock Encoding in style: a stylegan encoder for image-to-image translation.
\newblock In \emph{Proceedings of the IEEE/CVF conference on computer vision
  and pattern recognition}, 2287--2296.

\bibitem[{Sang et~al.(2022)Sang, Zhi, Song, Liu, Lai, Liu, Wen, Davis, and
  Luo}]{sang2022agileavatar}
Sang, S.; Zhi, T.; Song, G.; Liu, M.; Lai, C.; Liu, J.; Wen, X.; Davis, J.; and
  Luo, L. 2022.
\newblock AgileAvatar: Stylized 3D Avatar Creation via Cascaded Domain
  Bridging.
\newblock \emph{arXiv preprint arXiv:2211.07818}.

\bibitem[{Shi et~al.(2019)Shi, Yuan, Fan, Zou, Shi, and Liu}]{shi2019face}
Shi, T.; Yuan, Y.; Fan, C.; Zou, Z.; Shi, Z.; and Liu, Y. 2019.
\newblock Face-to-parameter translation for game character auto-creation.
\newblock In \emph{Proceedings of the IEEE/CVF International Conference on
  Computer Vision}, 161--170.

\bibitem[{Shi et~al.(2020)Shi, Zuo, Yuan, and Fan}]{shi2020fast}
Shi, T.; Zuo, Z.; Yuan, Y.; and Fan, C. 2020.
\newblock Fast and robust face-to-parameter translation for game character
  auto-creation.
\newblock In \emph{Proceedings of the AAAI Conference on Artificial
  Intelligence}, volume~34, 1733--1740.

\bibitem[{Shi et~al.(2022)Shi, Yang, Wan, and Shen}]{shi2022semanticstylegan}
Shi, Y.; Yang, X.; Wan, Y.; and Shen, X. 2022.
\newblock SemanticStyleGAN: Learning Compositional Generative Priors for
  Controllable Image Synthesis and Editing.
\newblock In \emph{Proceedings of the IEEE/CVF Conference on Computer Vision
  and Pattern Recognition}, 11254--11264.

\bibitem[{Song et~al.(2021)Song, Luo, Liu, Ma, Lai, Zheng, and
  Cham}]{song2021agilegan}
Song, G.; Luo, L.; Liu, J.; Ma, W.-C.; Lai, C.; Zheng, C.; and Cham, T.-J.
  2021.
\newblock AgileGAN: stylizing portraits by inversion-consistent transfer
  learning.
\newblock \emph{ACM Transactions on Graphics (TOG)}, 40(4): 1--13.

\bibitem[{Tuan~Tran et~al.(2017)Tuan~Tran, Hassner, Masi, and
  Medioni}]{tuan2017regressing}
Tuan~Tran, A.; Hassner, T.; Masi, I.; and Medioni, G. 2017.
\newblock Regressing robust and discriminative 3D morphable models with a very
  deep neural network.
\newblock In \emph{Proceedings of the IEEE conference on computer vision and
  pattern recognition}, 5163--5172.

\bibitem[{Wang et~al.(2019)Wang, Ma, Chen, Luo, Yi, and
  Bailey}]{wang2019symmetric}
Wang, Y.; Ma, X.; Chen, Z.; Luo, Y.; Yi, J.; and Bailey, J. 2019.
\newblock Symmetric cross entropy for robust learning with noisy labels.
\newblock In \emph{Proceedings of the IEEE/CVF International Conference on
  Computer Vision}, 322--330.

\bibitem[{Wolf, Taigman, and Polyak(2017)}]{wolf2017unsupervised}
Wolf, L.; Taigman, Y.; and Polyak, A. 2017.
\newblock Unsupervised creation of parameterized avatars.
\newblock In \emph{Proceedings of the IEEE International Conference on Computer
  Vision}, 1530--1538.

\bibitem[{Zhang et~al.(2018)Zhang, Isola, Efros, Shechtman, and
  Wang}]{zhang2018unreasonable}
Zhang, R.; Isola, P.; Efros, A.~A.; Shechtman, E.; and Wang, O. 2018.
\newblock The unreasonable effectiveness of deep features as a perceptual
  metric.
\newblock In \emph{Proceedings of the IEEE conference on computer vision and
  pattern recognition}, 586--595.

\bibitem[{Zhu et~al.(2020)Zhu, Shen, Zhao, and Zhou}]{zhu2020domain}
Zhu, J.; Shen, Y.; Zhao, D.; and Zhou, B. 2020.
\newblock In-domain gan inversion for real image editing.
\newblock In \emph{European conference on computer vision}, 592--608. Springer.

\end{thebibliography}

\section*{Appendix}
\subsection*{Implementation Details}
In the process of GAN-inversion, we optimize the latent code with initialization $w_{mean}$.
Notably, As glasses is an optional discrete avatar vector, normally using $w_{mean}$ may cause the absence of glasses when it should appear due to their transparent design in SemanticStyleGAN. Since the latent code of SemanticStyleGAN is disentangle to different local part, we select the latent codes which contains the glasses and average them to $w_{mean}^{glasses}$, then extract the glasses part and replace it into $w_{mean}$ to form a new mean latent code $\tilde{w}_{mean}$, it can generate glasses correctly after optimization and we use it in practical.

\begin{table}[h]
    \centering
    \begin{tabular}{l|l|l|l}
        \toprule
        Components     & Vectors        & Numbers     & Value Type      \\
        \midrule   
        head           & head type      & 6           & continuous  \\
        head           & skin tone      & 39          & discrete    \\
        head           & head width     & 1           & continuous  \\
        head           & head length    & 1           & continuous  \\
        \midrule
        eye            & eye type       & 4           & continuous  \\
        eye            & eye rotation   & 1           & continuous  \\
        eye            & eye spacing    & 1           & continuous  \\
        eye            & eye size       & 1           & continuous  \\
        \midrule
        mouth          & mouth type     & 9           & discrete    \\
        mouth          & mouth width    & 1           & continuous  \\
        mouth          & mouth volume   & 1           & continuous  \\
        mouth          & mouth position & 1           & continuous  \\
        \midrule
        nose           & nose type      & 5           & discrete    \\
        nose           & nose width     & 1           & continuous  \\
        nose           & nose height    & 1           & continuous  \\
        nose           & nose position  & 1           & continuous  \\
        \midrule
        hair           & hair type      & 36          & discrete    \\
        hair           & hair color     & 9           & discrete    \\
        \midrule
        brow           & type           & 8           & discrete    \\
        facial hair$^*$  & type           & 4           & discrete  \\
        glasses$^*$      & type           & 1           & discrete  \\
        \bottomrule
    \end{tabular}
    \caption{A detailed description of avatar vector in TikTok engine. (* indicates optional)}
    \label{tab:tiktok}
    \end{table}

\subsection*{Avatar Engine Information}
Here we provide detailed information of avatar engine we used in the experiments, which are shown in Table. \ref{tab:tiktok} and Table. \ref{tab:alter}. "Component" represents avatar parts which avatar vectors belong to. "Vectors" represents the parameter users can adjust (continuous) or select (discrete). "Numbers" of discrete vectors represents the assets numbers, "Number" of continuous vectors is 1. "Value Type" indicates whether continuous or discrete vectors.

The information of TikTok avatar engine is listed in Table. \ref{tab:tiktok}. It contains 8 discrete avatar vectors and 13 continuous vectors. 
And Alter avatar engine only contains 6 discrete avatar vectors, which is listed in Table. \ref{tab:alter}. The experiments on two diverse avatar engines demonstrates the generality of our method.

\begin{table}[h]
    \centering
    \begin{tabular}{l|l|l|l}
        \toprule
        Components     & Controllers    & numbers     & value type      \\
        \midrule   
        head           & head type      & 6           & discrete    \\
        head           & skin tone      & 8          & discrete    \\
        \midrule
        hair           & hair type      & 22          & discrete    \\
        hair           & hair color     & 7           & discrete    \\
        \midrule
        brow           & type           & 5           & discrete    \\
        glasses$^*$      & type           & 1           & discrete  \\
        \bottomrule
    \end{tabular}
    \caption{A detailed description of avatar vector in Alter engine. (* indicates optional)}
    \label{tab:alter}
    \end{table}

\subsection*{More Visual Results}
\begin{figure*}[h]
\centering
\includegraphics[width=0.95\textwidth]{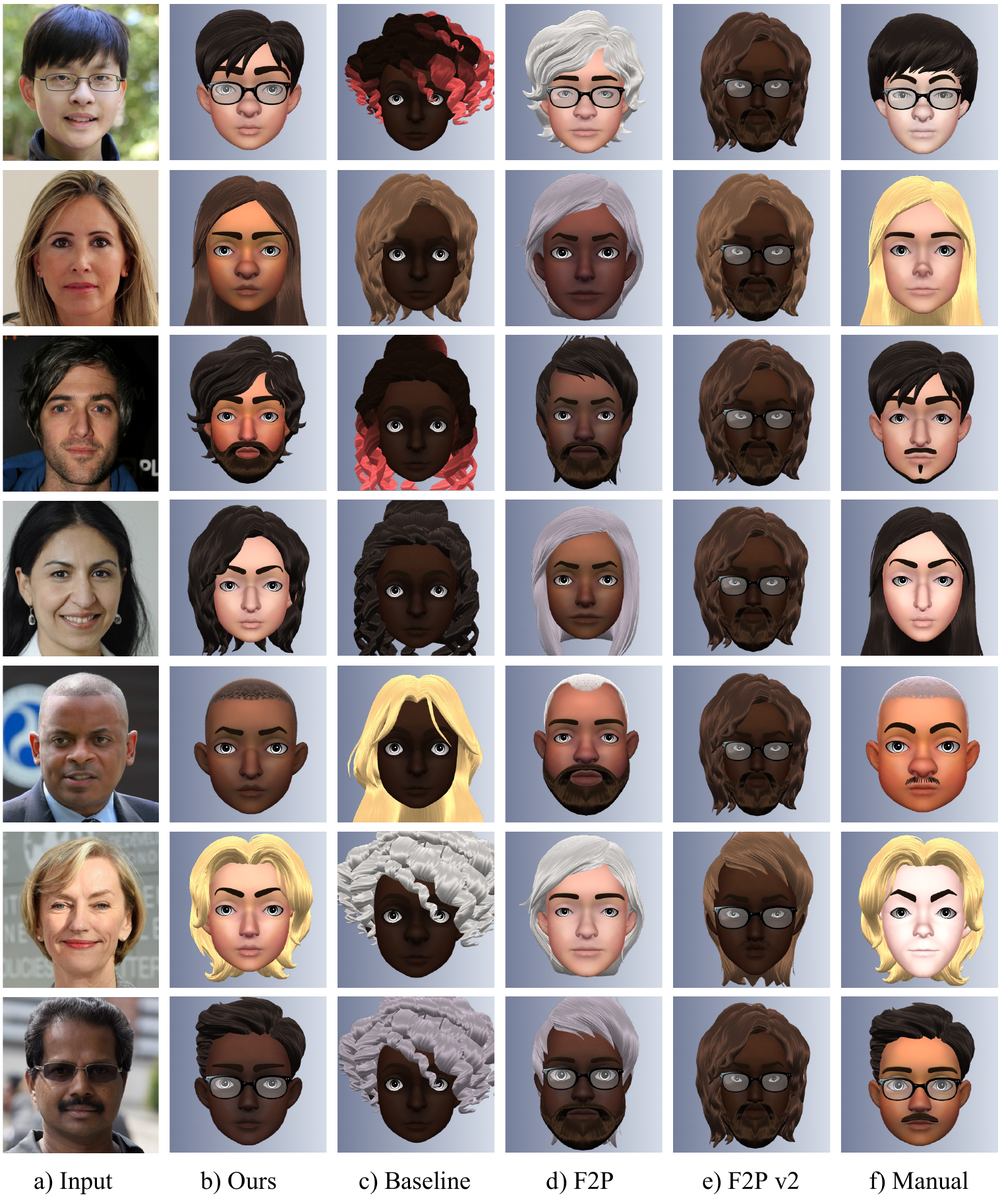}
\caption{More results on TikTok  engine: Input, Ours, Baseline, F2P, F2P v2, Manual}
\label{fig:supp-tiktok}
\end{figure*}

\begin{figure*}[h]
\centering
\includegraphics[width=0.95\textwidth]{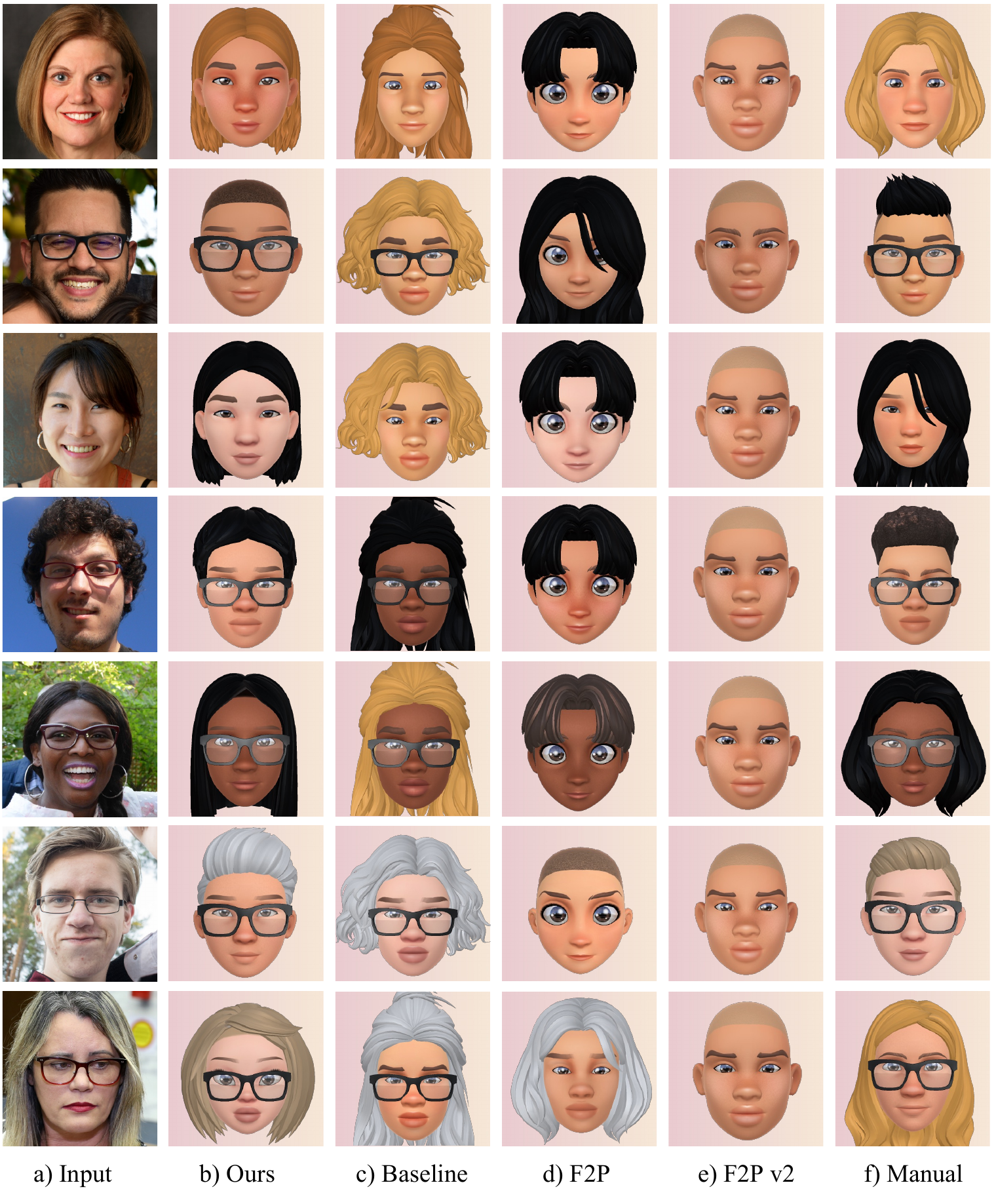}
\caption{More results on Alter engine: Input, Ours, Baseline, F2P, F2P v2, Manual}
\label{fig:supp-alter}
\end{figure*}

Figure. \ref{fig:supp-tiktok} and Figure. \ref{fig:supp-alter} show more visual comparison on TikTok engine and Alter engine.

\clearpage

\end{document}